\documentclass[sigconf]{acmart}

\usepackage{bm}
\usepackage{amssymb}
\usepackage{multirow}
\usepackage{subfigure}
\usepackage{array}
\usepackage{graphicx}
\usepackage{amsmath}
\usepackage{threeparttable}
\usepackage{dsfont}

\newcommand{\figref}[1]{Figure \ref{#1}}
\newcommand{\tabref}[1]{Table \ref{#1}}
\newcommand{\secref}[1]{Section \ref{#1}}
\newcommand{\equref}[1]{Equation (\ref{#1})}

\copyrightyear{2020}
\acmYear{2020}
\setcopyright{acmcopyright}
\acmConference[KDD '20]{Proceedings of the 26th ACM SIGKDD Conference on Knowledge Discovery and Data Mining}{August 23--27, 2020}{Virtual Event, CA, USA}
\acmBooktitle{Proceedings of the 26th ACM SIGKDD Conference on Knowledge Discovery and Data Mining (KDD '20), August 23--27, 2020, Virtual Event, CA, USA} 
\acmPrice{15.00}
\acmDOI{10.1145/3394486.3403152}
\acmISBN{978-1-4503-7998-4/20/08}

\begin{document}
\fancyhead{}

\title[
Predicting Temporal Sets with Deep Neural Networks
]{
Predicting Temporal Sets with Deep Neural Networks
}
\author{Le Yu$^1$, Leilei Sun$^{1\ast}$, Bowen Du$^{1}$, Chuanren Liu$^{2}$, Hui Xiong$^3$, Weifeng Lv$^1$}\thanks{$\ast$ Corresponding author}

\affiliation{
\textsuperscript{1}{SKLSDE and BDBC Lab, Beihang University, Beijing 100083, China}
\\
\textsuperscript{2}{Department of Business Analytics and Statistics, University of Tennessee, Knoxville, USA}
\\
\textsuperscript{3}{Department of Management Science and Information Systems, Rutgers University, USA}
}

\affiliation{\textsuperscript{1}{\{yule,leileisun,dubowen,lwf\}@buaa.edu.cn},  \textsuperscript{2}{cliu89@utk.edu},  \textsuperscript{3}{hxiong@rutgers.edu}}


\begin{abstract}
Given a sequence of sets, where each set contains an arbitrary number of elements, the problem of temporal sets prediction aims to predict the elements in the subsequent set.
In practice, temporal sets prediction is much more complex than predictive modelling of temporal events and time series, and is still an open problem.
Many possible existing methods, if adapted for the problem of temporal sets prediction, usually follow a two-step strategy by first projecting temporal sets into latent representations and then learning a predictive model with the latent representations.
The two-step approach often leads to information loss and unsatisfactory prediction performance.
In this paper, we propose an integrated solution based on the deep neural networks for temporal sets prediction.
A unique perspective of our approach is to learn element relationship by constructing set-level co-occurrence graph and then perform graph convolutions on the dynamic relationship graphs.
Moreover, we design an attention-based module to adaptively learn the temporal dependency of elements and sets.
Finally, we provide a gated updating mechanism to find the hidden shared patterns in different sequences and fuse both static and dynamic information to improve the prediction performance.
Experiments on real-world data sets demonstrate that our approach can achieve competitive performances even with a portion of the training data and can outperform existing methods with a significant margin.
\end{abstract}




\maketitle

\section{Introduction}\label{section-1}
\begin{figure}[!htbp]
    \centering
    \includegraphics[width=\columnwidth]{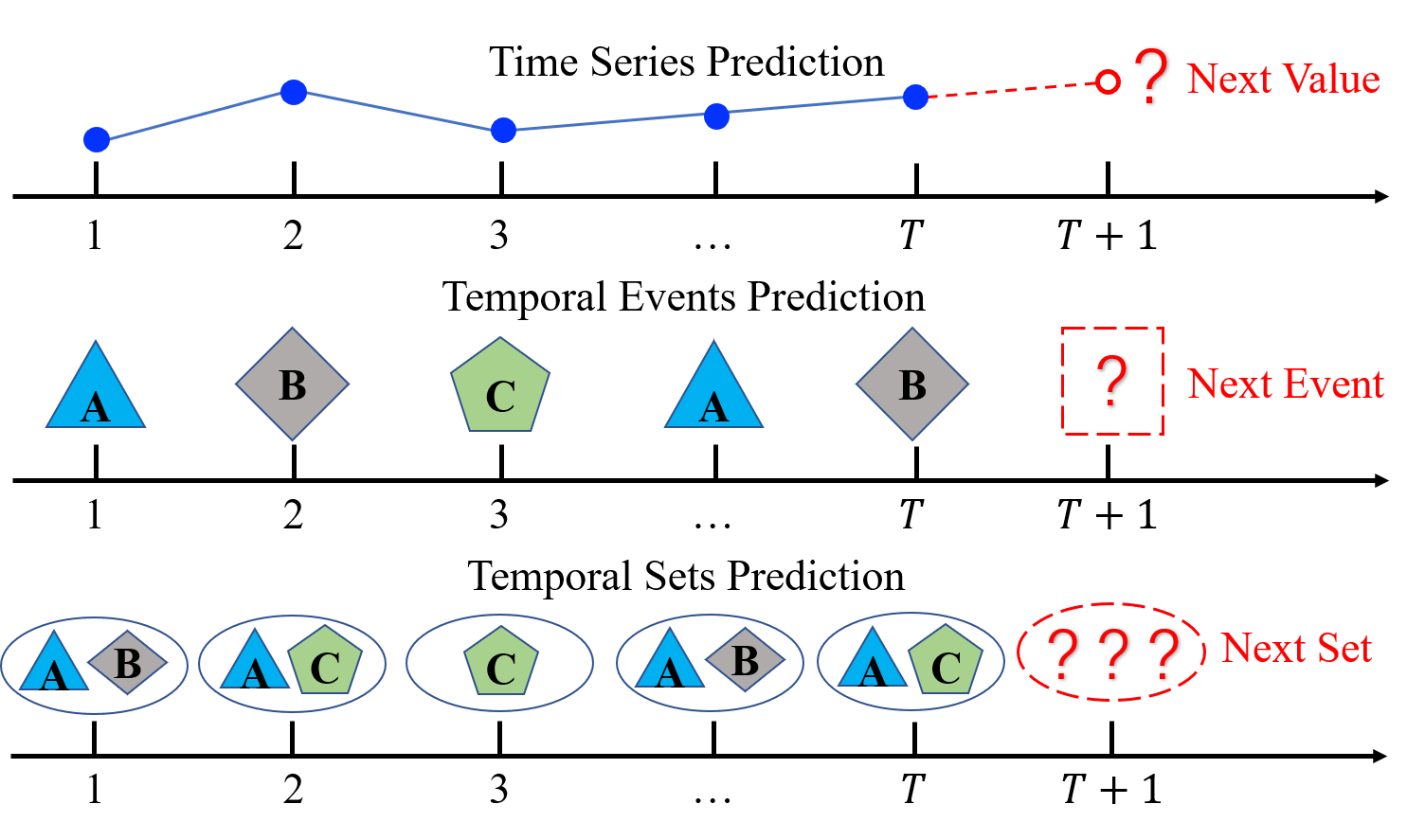}
    \caption{Prediction of three types of temporal data: time series, temporal events and temporal sets.}
    \label{fig:temporal_data}
\end{figure}

Temporal data record the objects vary over time and the time-oriented nature of such data makes them valuable.
Mining the underlying patterns and dynamics in temporal data could help people make better decisions or plans. 
For example, forecasting the speed of traffic flow provides better strategies for transportation departments \cite{DBLP:conf/iclr/LiYS018}. 
Predicting the labor mobility contributes to human resource reallocation in labor markets \cite{li2017nemo}. 
Due to the importance of temporal data, a great number of temporal data mining methods have been proposed \cite{laxman2006survey,DBLP:journals/tkde/GuptaGAH14}.
However, most of the existing methods were designed for time series \cite{brockwell2002introduction,che2018recurrent} or temporal events \cite{sun2016data,li2017nemo}.
This paper studies the prediction of a new type of temporal data, namely, temporal sets \cite{DBLP:conf/kdd/Benson0T18}.
If time series could be seen as a sequence of numerical values recorded with timestamps, temporal events could be seen as a sequence of nominal events with timestamps, and then temporal sets are a sequence of sets with timestamps, where each set contains an arbitrary number of elements, see \figref{fig:temporal_data}.

In fact, temporal sets are very pervasive in real-world scenarios. For example, a customer's purchase behaviors could be formalized as a sequence of sets, where each set includes a number of goods and corresponds to a purchase at a supermarket. Forecasting student's next-semester courses selection \cite{xu2016personalized} and predicting patient's next-period prescriptions \cite{wang2018supervised,jin2018treatment} also deal with this type of temporal data.
It is no doubt that temporal sets prediction is of great importance. Take the above scenarios as instances, prediction of next-period basket could help stores dispatch products in advance, and predicting next-semester courses could help universities make better decisions about course setting. However, the existing temporal data prediction method designed for time series or temporal events could not be directly used for temporal sets because time series prediction method can not handle semantic relationships among elements, while temporal events prediction method cannot deal with multiple elements within a set.

Recent literature has reported a few methods for temporal sets prediction \cite{yu2016dynamic,choi2016multi,DBLP:conf/kdd/HuH19}. These methods were designed under a two-stage framework, which first projected each set into a latent vector and then predicted the subsequent set based on the sequences of embedded sets. \citet{choi2016multi} introduced a variant of Skip-gram model to learn the representations of sets according to the co-occurrence of elements, and a softmax classifier was utilized to predict the subsequent set within a context window. \citet{yu2016dynamic} and \citet{DBLP:conf/kdd/HuH19} first embedded sets into structured vectors by pooling operations and then learned dynamic hidden patterns in sequential behaviors by Recurrent Neural Networks (RNNs). However, the two-step methods suffered from information loss during the set representation process, which resulted in unsatisfactory prediction performance. Although in recent years, a lot of works on representation learning of set-based data have been proposed \cite{zaheer2017deep,DBLP:conf/icml/LeeLKKCT19}, the learned representations were mainly applied to downstream tasks, which did not take the dynamic sequential behaviors into consideration. Hence, for the task of temporal sets prediction, it is difficult to learn latent representations of sets and then mine sequential patterns based on the learned representations.

To address the above issues, we propose a novel \textbf{D}eep \textbf{N}eural \textbf{N}etwork for \textbf{T}emporal \textbf{S}ets \textbf{P}rediction, namely DNNTSP, which consists of three components: element relationship learning, attention-based temporal dependency learning and gated information fusing. In the proposed model, we consider the interactions of elements not only in the same set, but also among different sets. 
Different from the existing temporal sets prediction methods, we first propose a weighted graph convolutional network on dynamic graphs which aims to learn the relationships among elements in each set by information propagation. 
Then for the sequence of each appearing element, an attention-based module is designed to learn temporal dependency among different sets and aggregate historical hidden states into a latent vector. 
Finally, a gated updating mechanism is provided to fuse both the static and dynamic information of elements, which could achieve high prediction performance according to the comprehensive information it integrates.
In summary, this paper has the following contributions:
\begin{itemize}
    \item
    Different from the existing research which turns the temporal sets prediction problem into a conventional predictive modelling problem by a set embedding procedure, our method founds on comprehensive element representation which first captures element relationship by constructing set-level co-occurrence graph and then performs graph convolutions on the dynamic relationship graphs.
    
    \item
    An attention-based temporal dependency learning module is provided, which is able to capture the most important temporal dependencies among elements in the historical sequence of sets and then aggregate the temporal information by a weighted summation adaptively.
    
    \item
    A gated updating mechanism is designed to fuse both the static and dynamic representations of elements, which improves the prediction performance by mining the shared dynamic temporal patterns among elements.
    
\end{itemize}


\section{Problem Formalization}\label{section-2}
This section first presents the definition of the temporal sets and then provides formalization of the studied problem. 
\begin{definition}
    \textbf{Temporal Sets}: Temporal sets can be treated as a sequence of sets, where each set consists of an arbitrary number of elements and also a timestamp.
\end{definition}
It is worth noting that temporal sets are quite pervasive in practice, because in many real-world scenarios, a number of individual or group behaviors are recorded with a same time label. For example, purchasing a collection of goods at a visit to a supermarket, selecting a number of courses at a semester, etc. As a new category of temporal data, temporal sets are more complicated than time series and temporal events. 
 
The studied problem of this paper, temporal sets prediction, could be formalized as follows. Let $\mathbb{U}=\{u_1,\cdots,u_n\}$ and $\mathbb{V}=\{v_1,\cdots,v_m\}$ be the collections of $n$ users and $m$ elements respectively.
A set $S$ is a collection of elements, $S \subset \mathbb{V}$.
Given a sequence of sets $\mathbb{S}_i=\left\{S_i^1,S_i^2,\cdots,S_i^T\right\}$ that records the historical behaviors of user $u_i \in \mathbb{U}$, the goal of temporal sets prediction is to predict the subsequent set according to the historical records, that is,
$$\hat{S}_i^{T+1} = f(S_i^1,S_i^2,\cdots,S_i^T,\bm{W}),$$
where $\bm{W}$ represents the trainable parameters. To solve the above problem, one needs to consider 
the relationships among elements and the temporal dependency underlying the set sequence. 
Therefore, existing temporal data prediction methods for time series and temporal events cannot be applied to temporal sets directly.

\section{Methodology}\label{section-3}
This section first presents the framework of the proposed model and then introduces the components step by step.

\begin{figure}[!htbp]
    \centering
    \includegraphics[scale = 0.36]{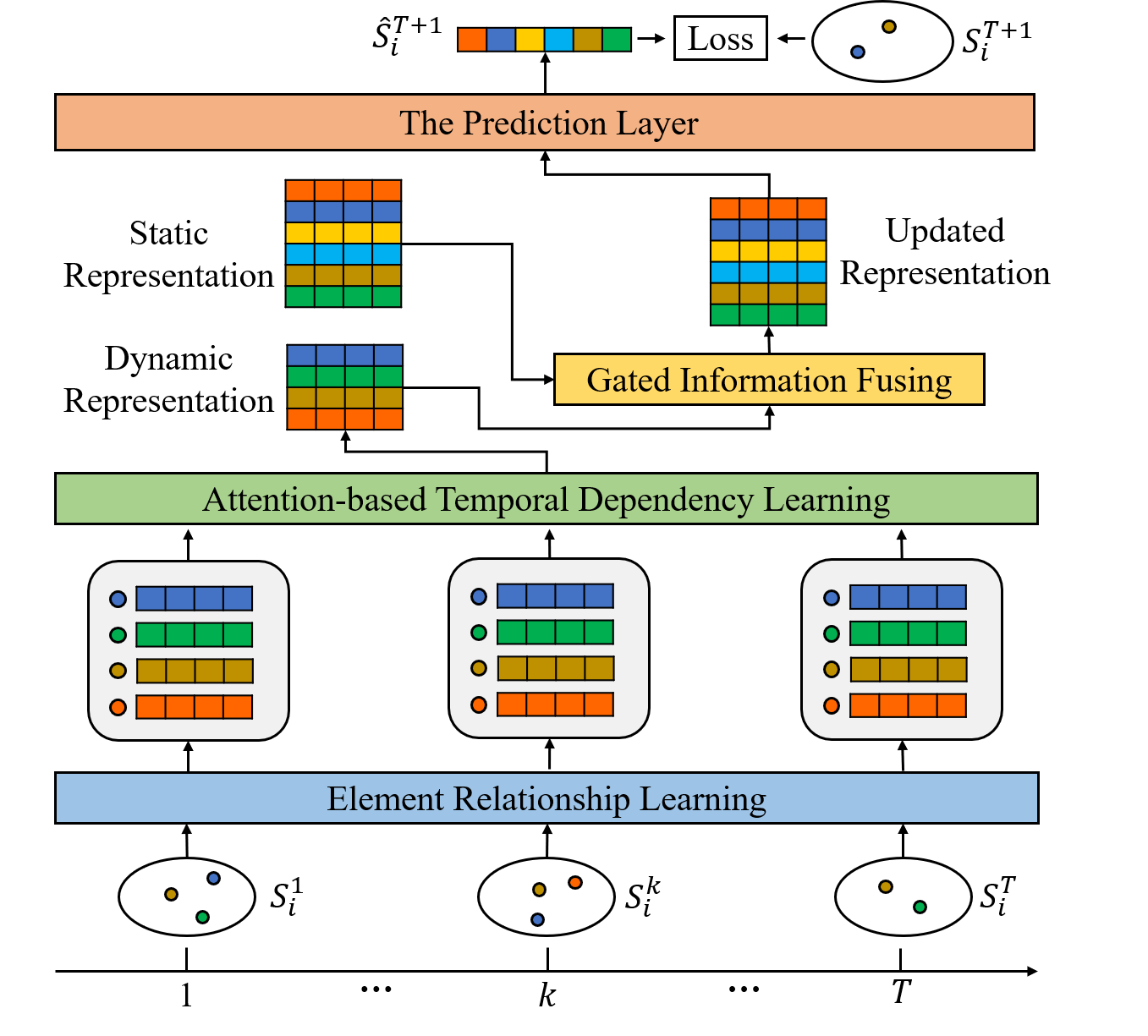}    
    \caption{Framework of the proposed model.}
    \label{fig:framework}
\end{figure}

The framework of the proposed model is shown in \figref{fig:framework}, which consists of three components: element relationship learning, attention-based temporal dependency learning and gated information fusing.
The first component is designed to learn set-level element relationship, which first constructs weighted graphs based on the co-occurrence of elements, then propagates information among elements on dynamic graphs, and finally obtains each element's updated representation based on the received information. 
The second component aims to learn temporal dependency of each element in different sets. This component first takes the sequence of element's representations as the input and uses the attention mechanism to learn the temporal dependencies of sets and elements from the past sequences. Then it provides an aggregation of historical states of elements to the next component.
The third component assumes that the interactions of elements could be shared among different sequences. It fuses static and dynamic representations together by a gated updating mechanism, which could improve the final prediction performance by considering all the collected information comprehensively.

\subsection{Element Relationship Learning}
Most of the existing temporal sets prediction methods have two main components: set embedding and temporal dependency learning, where set embedding turns the temporal sets prediction problem into a conventional predictive modelling problem. However, the set embedding step suffers from information loss problem caused by the pooling operation, which reduces the final prediction accuracy.

In order to leverage the useful information in element relationship as much as possible, this paper founds on element representation according to set-level relationship learning. In particular, we propose a weighted graph convolutional network on dynamic graphs, which first constructs weighted graphs based on the co-occurrence of elements and then propagates information between elements in each graph. Let $\bm{E} \in \mathbb{R}^{m \times F}$ denote the embedding matrix of all elements \footnote{The embedding matrix $\bm{E}$ is initialized from the standard normal distribution.}, where $F$ is the dimension of element representation. Then we learn the relationships of elements by the following two steps.

\begin{figure}[!htbp]
    \centering
    \includegraphics[width=0.96\columnwidth]{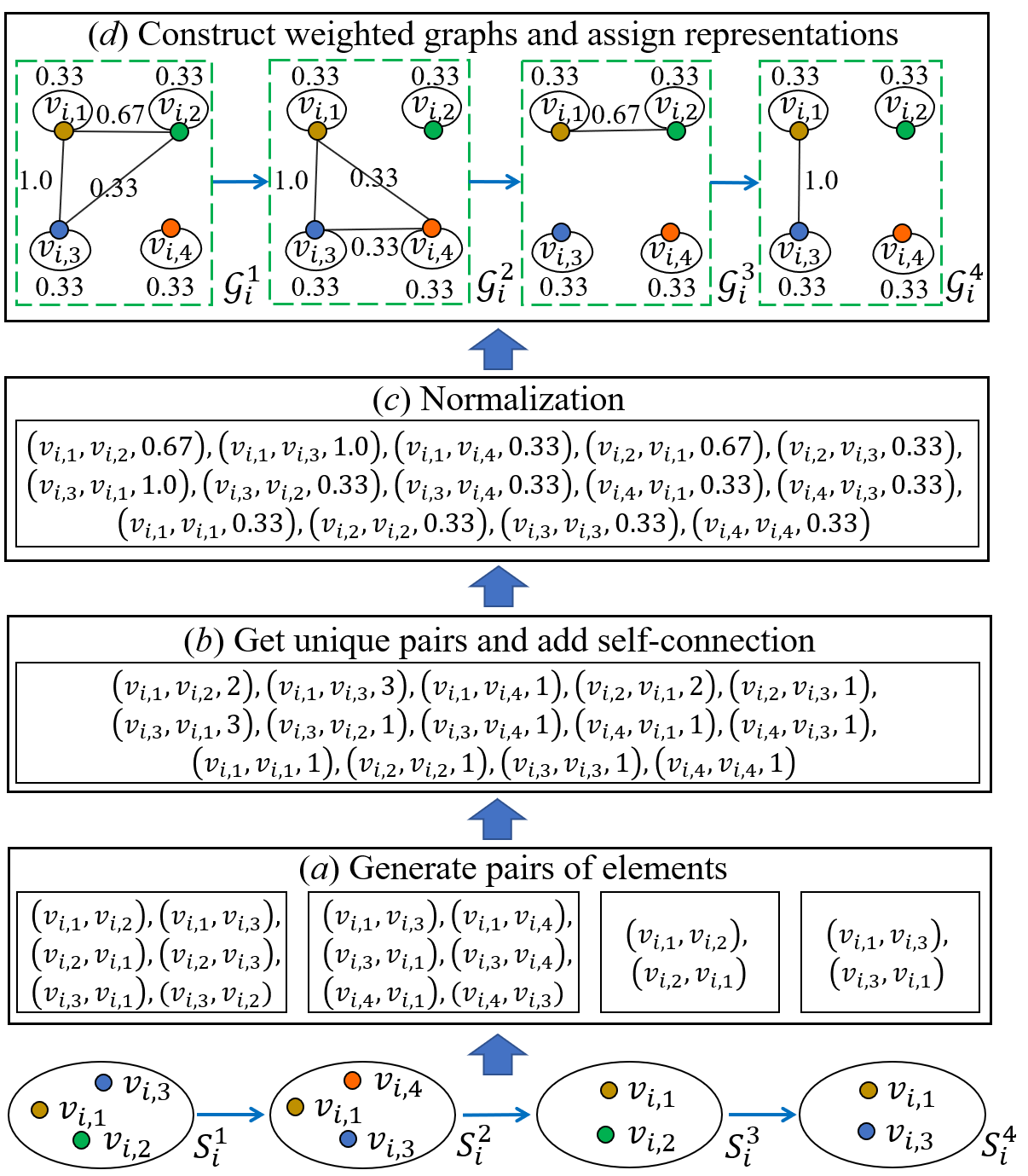}
    \caption{The process of weighted graphs construction.}
    \label{fig:weighted_graph_construction}
\end{figure}

\textbf{Weighted Graphs Construction}.
The process of constructing weighted graphs is shown in \figref{fig:weighted_graph_construction}.
For $S_i^t \in \mathbb{S}_i$, the $t$-th set of user $u_i$, we first generate the pairs of every two elements in $S_i^t$ and each pair denotes a co-occurrence relationship of two elements in $S_i^t$. For example, let $S_i^1=\{v_{i,1},v_{i,2},v_{i,3}\}$ and the generated pairs are $(v_{i,1}, v_{i,2})$, $(v_{i,2}, v_{i,1})$, $(v_{i,1}, v_{i,3})$, $(v_{i,3}, v_{i,1})$, $(v_{i,2}, v_{i,3})$ and $(v_{i,3}, v_{i,2})$, see \figref{fig:weighted_graph_construction}(\textit{a}). After generating pairs of elements for each set in $\mathbb{S}_i$, we could obtain a collection of all pairs. Then we select all unique pairs and assign value to each pair based on its appearing frequency. We add self-connection for each element appearing in $\mathbb{S}_i$ by treating its appearing frequency as 1, which is used to reduce the information loss for rarely appearing elements in the sequence as the information of such elements would decrease dramatically in long sequences without the self-connection during the following convolutional operations, see \figref{fig:weighted_graph_construction}(\textit{b}). After that, we normalize the value of each pair between 0 and 1 to denote the weights among different elements as shown in \figref{fig:weighted_graph_construction}(\textit{c}). Finally, we construct the graph for each set based on the calculated weights and assign representation to each element by its corresponding representation in $\bm{E}$, see \figref{fig:weighted_graph_construction}(\textit{d}). Following the above steps, we could construct $T$ weighted dynamic graphs $\mathbb{G}_i=\left\{\mathcal{G}_i^1,\mathcal{G}_i^2,\cdots,\mathcal{G}_i^T\right\}$. The $t$-th graph $\mathcal{G}_i^t=(\mathcal{V}_i,\mathcal{E}_i^t)$ is a weighted undirected graph with a weighted matrix $\bm{A}_i^t \in \mathbb{R}^{|\mathcal{V}_i| \times |\mathcal{V}_i|}$, where $\mathcal{V}_i$ denotes the set of appearing elements in $\mathbb{S}_i$ and $\mathcal{E}_i^t$ denotes the set of edges in $\mathcal{G}_i^t$. In the following parts, we use $\bm{e}_{i,j}^t \in \mathbb{R}^F$ to denote the representation of element $v_{i,j} \in \mathcal{V}_i$ at time $t$.

\textbf{Weighted Convolutions on Dynamic Graphs}.
This paper designs a novel module to perform weighted convolutions on the constructed dynamic graphs. The input of this module is a sequence of dynamic graphs $\mathbb{G}_i=\left\{\mathcal{G}_i^1,\mathcal{G}_i^2,\cdots,\mathcal{G}_i^T\right\}$, where graph $\mathcal{G}_i^t \in \mathbb{G}_i$ has a sequence of elements represented as $\left\{\bm{e}_{i,j}^t \in \mathbb{R}^F, \forall v_{i,j} \in \mathcal{V}_i\right\}$. For graph $\mathcal{G}_i^t$, the output of this module is a new sequence of element representation, which could be denoted as $\left\{\bm{c}_{i,j}^t \in \mathbb{R}^{F^\prime}, \forall v_{i,j} \in \mathcal{V}_i\right\}$, where each element is denoted with $F^\prime$ dimensions.

The weighted convolutions are implemented by propagating information of elements in each dynamic graph as follows. Take graph $\mathcal{G}_i^t$ as an instance,
\begin{equation}
  \label{equ:origin_equ}
  \bm{c}_{i,j}^{t,l+1}=\sigma\left(\bm{b}^{t,l}+\sum_{k \in \mathcal{N}_{i,j}^t \cup \{j\}} \bm{A}_i^{t}[j,k]\cdot\left(\bm{W}^{t,l}\bm{c}_{i,k}^{t,l}\right)\right), 
\end{equation}
where $\bm{A}_i^{t}[j,k]$ represents the item at the $j$-th row and $k$-th column of matrix $\bm{A}_i^t$, which is the edge weight of $v_{i,j}$ and $v_{i,k}$ in graph $\mathcal{G}_i^t$, $\bm{W}^{t,l} \in \mathbb{R}^{ F^l \times F^{l-1}}$ and $\bm{b}^{t,l} \in \mathbb{R}^{F^l}$ are trainable parameters of the $l$-th convolutional layer at time $t$, and $\bm{c}_{i,j}^{t,l}$ denotes the representation of $v_{i,j} \in \mathcal{V}_i$ in the $l$-th layer at time $t$. $F^l$ denotes the output dimension of the $l$-th layer and $F^0$ is equal to $F$. $\mathcal{N}_{i,j}^t$ are neighbors' indices of the $j$-th element in graph $\mathcal{G}_i^t$. To reduce the parameter scale and also make our method flexible to deal with sequences with variable lengths, a parameter sharing strategy is adopted, Equation (1) is rewritten as
\begin{equation}
  \label{equ:share_equ}
  \bm{c}_{i,j}^{t,l+1}=\sigma\left(\bm{b}^{l}+\sum_{k \in \mathcal{N}_{i,j}^t \cup \{j\}} \bm{A}_i^{t}[j,k]\cdot\left(\bm{W}^{l}\bm{c}_{i,k}^{t,l}\right)\right),
\end{equation}
which means that we utilize shared parameters for convolutional layers across different timestamps. In the first layer, $\bm{c}_{i,j}^{t,0}$ is initialized from the standard normal distribution, which is actually the representation of $v_{i,j}$ in $\bm{E}$. The output dimension of the last layer is set to $F^\prime$. Due to the weighted convolutions on dynamic graphs, each element in the graphs could not only receive the information from itself, but also receive the information from its neighbours. The representation of each element is updated by all the received information. After the information propagating thoroughly, we achieve a stable representation of each element, which comprehensively considers the relationships of all elements in the graphs. Formally, we use $C_{i,j}=\left\{\bm{c}_{i,j}^1,\bm{c}_{i,j}^2,\cdots,\bm{c}_{i,j}^T\right\}$ to denote the sequence of $v_{i,j}$, where $\bm{c}_{i,j}^t \in \mathbb{R}^{F^\prime}$ is the output of the last convolutional layer.

\subsection{Attention-based Temporal Dependency Learning}
For the studied problem, it has been reported that some elements appear quite frequently and regularly in a sequence, while the other elements appear irregularly and occasionally \cite{DBLP:conf/kdd/HuH19}, which makes the temporal dependency among a sequence dynamic and complicated. Traditional RNNs fail to handle such temporal dependency, even some gated model have been proposed (e.g. LSTM \cite{DBLP:journals/neco/HochreiterS97}, GRU \cite{chung2014empirical}). The reason is that RNNs only propagate information sequentially, which limits the perception field of temporal dependency learning \cite{DBLP:conf/acl/JurafskyHQK18}. Different from the RNNs, the self-attention mechanism could provide a model with the ability to capture the temporal dependency without such limitation \cite{vaswani2017attention,sankar2018dynamic}. In our model, we extend the self-attention mechanism to capture temporal dependency.

To learn the dynamic and evolutionary patterns in sequences, a temporal dependency learning component is proposed in this paper. The inputs of this component are the sequences of all elements' representations in $\mathcal{V}_i$, which could be denoted as $\mathbb{C}_i=\left\{C_{i,1},C_{i,2},\cdots,C_{i,|\mathcal{V}_i|}\right\}$, where $C_{i,j}=\left\{\bm{c}_{i,j}^1,\bm{c}_{i,j}^2,\cdots,\bm{c}_{i,j}^T\right\}$ are the representations of element $v_{i,j}$ over time. We use $\bm{C}_{i,j} \in \mathbb{R}^{T \times F^\prime}$ to denote the stacked matrix representation of $C_{i,j}$, where the $t$-th row of $\bm{C}_{i,j}$ is $\bm{c}_{i,j}^t$. The outputs of this component are aggregated and compact representations of elements in $\mathcal{V}_i$, that is, $\mathbb{Z}_i=\{\bm{z}_{i,1},\bm{z}_{i,2},\cdots,\bm{z}_{i,|\mathcal{V}_i|}\}$, where $\bm{z}_{i,j} \in \mathbb{R}^{F^{\prime\prime}}$ denotes the representation of element $v_{i,j} \in \mathcal{V}_i$. This subsection first introduces how to aggregate the stacked representations $\bm{C}_{i,j}$ into new representations $\bm{Z}_{i,j}$ with consideration of temporal dependency, and then discusses how to compress the new representations into a compact representation of $v_{i,j}$, denoted as $\bm{z}_{i,j}$.

The self-attention is used to learn temporal dependency of the stacked representations for each element. 
The new representations with consideration of temporal dependency are computed as
\begin{equation}
    \bm{Z}_{i,j}=softmax\left(\frac{(\bm{C}_{i,j}\bm{W}_q)\cdot(\bm{C}_{i,j}\bm{W}_k)^\top}{\sqrt{F^{\prime\prime}}}+\bm{M}_i\right)\cdot\left(\bm{C}_{i,j}\bm{W}_v\right),
\end{equation}
where $\bm{W}_q \in \mathbb{R}^{F^\prime \times F^{\prime\prime}}$, $\bm{W}_k \in \mathbb{R}^{F^\prime \times F^{\prime\prime}}$, $\bm{W}_v \in \mathbb{R}^{F^\prime \times F^{\prime\prime}}$ are trainable parameters to calculate queries, keys and values for elements in the sequence, $\bm{Z}_{i,j} \in \mathbb{R}^{T \times F^{\prime\prime}}$ is the stacked representation of $v_{i,j}$'s sequence. $\bm{M}_i \in \mathbb{R}^{T \times T}$ is a masked matrix, which is used to avoid the future information leakage and guarantee that the state of each timestamp is only affected by its previous states. It is defined as
\begin{equation}
    \notag
    M_i^{t,t^\prime}=
    \begin{cases}
    0 & \text{if } t \leq t^\prime,\\
    -\infty & \text{otherwise}.
    \end{cases} 
\end{equation}
Then we aggregate the sequential information into a vectorized representation by the following weighted aggregation equation,
\begin{equation}
    \bm{z}_{i,j}=\left(\left(\bm{Z}_{i,j}\cdot\bm{w}_{agg}\right)^\top\cdot\bm{Z}_{i,j}\right)^\top,
\end{equation}
where $\bm{w}_{agg} \in \mathbb{R}^{F^{\prime\prime}}$ is a trainable parameter to learn the importance of different timestamps adaptively. $\bm{z}_{i,j} \in \mathbb{R}^{F^{\prime\prime}}$ is a compact representation for element $v_{i,j}$ that considers all the possible temporal dependencies.
In this paper, we set $F^{\prime\prime}=F$ to make the calculation in the following part more convenient.

\subsection{Gated Information Fusing}
Our model predicts the subsequent set based solely on the historical behaviors, without using any other auxiliary information (e.g. the attributes of users). 
This indicates that different users may share the same patterns in their sequential behaviors. Mining the shared patterns could not only make our method suitable for sparse data but also improve the robustness of the prediction result.

To discover the shared hidden patterns and also to combine the static and dynamic information together, A gated information fusing component is provided here. 
The input of this component has two parts: 
the shared element representation matrix $\bm{E}$ and the compact representations of elements w.r.t. user $u_i$, $\mathbb{Z}_i=\{\bm{z}_{i,1},\bm{z}_{i,2},\cdots,\bm{z}_{i,|\mathcal{V}_i|}\}$. 
The first part $\bm{E}$ could be treated as the static representations of elements as it is shared by all the users. 
The second part $\mathbb{Z}_i$ could be seen as the dynamic representations of elements appearing in $\mathcal{V}_i$ because it considers both the co-occurrence relationships and the temporal dependency of the elements. 
We use $\bm{E}_i$ to denote the hidden state of user $i$, which is initialized as $\bm{E}$.
The most recent state $\bm{E}_{i,I(j)}^{update}$ is achieved by updating the user state $\bm{E}_i$ iteratively as follows,
\begin{equation}
    \label{equ:gated_update}
    \begin{aligned}
        \bm{E}_{i,I(j)}^{update}&=(1-\beta_{i,I(j)} \cdot \gamma_{I(j)})\cdot\bm{E}_{i,I(j)}+ (\beta_{i,I(j)} \cdot \gamma_{I(j)})\cdot\bm{z}_{i,j},
    \end{aligned}
\end{equation}
where $I(\cdot)$ is a function that maps element $v_{i,j}$ to its corresponding index in $\bm{E}_i$, $\beta_{i,j}$ and $\gamma_{j}$ are the $j$-th dimension of $\bm{\beta}_i$ and $\bm{\gamma}$. $\bm{\beta}_i \in \mathbb{R}^m$ is a indicator vector composed of 0 or 1, where the entry with value 1 means the corresponding element is in $\mathcal{V}_i$. $\bm{\gamma} \in \mathbb{R}^m$ is the trainable parameter of an updating gate which controls the importance of the static and dynamic representations. In \equref{equ:gated_update}, the representations of elements appearing in $\mathcal{V}_i$ are updated according to both the static and dynamic information. For the other elements, we just maintain its original static representations. 

\subsection{The Prediction Layer}
Finally, the possibilities of all elements appearing in the next-period set could be computed according to the user's current state,
\begin{equation}
    \hat{\bm{y}}_i=sigmoid(\bm{E}_i^{update} \cdot \bm{w}_o+b_o),
\end{equation}
where $\bm{w}_o \in \mathbb{R}^F$ and $b_o \in \mathbb{R}$ are trainable parameters to provide the final prediction result.

\subsection{The Learning Process}
To construct our temporal sets prediction model, we first stack multiple weighted convolutional layers with shared parameters and propagate information of elements on the dynamic graphs to learn set-level element relationship. 
Then we provide an attention-based temporal dependency learning module to learn the temporal dependency with a complete receptive field, and aggregate the temporal information into a latent representation for each element using the weighted aggregation. 
In order to enhance the learning capacity of this component, we use multiple heads to identify the most influencing modes, and concatenate the representations of different heads to get a joint representation. 
Moreover, we employ the gated information fusing component to take in the output of temporal dependency learning module and the embedding matrix to explore the shared patterns in different sequences. Finally, the prediction layer provides the final result.

In the training process, predicting next-period set could be treated as a multi-label learning problem, so we adopt the loss function with $L_2$ regularization technique as follows,
\begin{equation}
    \label{equ:loss_function}
    L=-\frac{1}{N}\sum_i^N\frac{1}{m}\sum_j^m {y_{i,j}\log(\hat{y}_{i,j})+(1-y_{i,j})\log(1-\hat{y}_{i,j})}+\lambda\|\bm{W}\|^2,
\end{equation}
where $\bm{W}$ denotes all the trainable parameters in our model, $N$ is the number of training samples, $\lambda$ is a hyperparameter to control the importance of $L_2$ regularization, $y_{i,j}$ and $\hat{y}_{i,j}$ denote the $j$-th dimension of the ground truth and the predicted appearing possibility of $j$-th element in the next set of user $u_i$. We optimize the proposed model by minimizing \equref{equ:loss_function} until convergence.

\section{Experiments}\label{section-4}
This section evaluates the performance of the proposed method by experiments on real-world data sets. Both classical and state-of-the-art methods are implemented to provide baseline performance, and multiple metrics are used to provide comprehensive evaluation.

\subsection{Description of Datasets}
We conduct experiments on four real-world datesets:
\begin{itemize}
	\item \textbf{TaFeng} \footnote{https://www.kaggle.com/chiranjivdas09/ta-feng-grocery-dataset}:
    TaFeng is a public dataset which contains four months of shopping transactions at a Chinese grocery store. This dataset was recorded at day-level, and we treat products purchased in the same day by the same customer as a set.
    
    \item \textbf{Dunnhumby-Carbo (DC)} \footnote{https://www.dunnhumby.com/careers/engineering/sourcefiles}: 
    DC contains two years of transactions from households at a retailer which could be available online. Products purchased in the same transaction are treated as a set. We select the transactions during the first two months to conduct experiments because it's costly to train on the original dataset due to its large scale.
    
    \item \textbf{TaoBao} \footnote{https://tianchi.aliyun.com/dataset/dataDetail?dataId=649}: 
    TaoBao contains lots of users who have four types of behaviors including clicking, purchasing, adding products to shopping carts and marking products as favor online. We select all purchasing behaviors and treat products bought in the same transaction as a set.
    
    \item \textbf{Tags-Math-Sx (TMS)} \footnote{https://math.stackexchange.com}: 
    TMS dataset contains the whole history of users in Mathematics Stack Exchange, a stack exchange for mathematics questions. We use the TMS dataset preprocessed by \citet{DBLP:conf/kdd/Benson0T18} to do experiments.
\end{itemize}
For simplicity, we select frequent elements which cover 80\% records in each dataset to conduct experiments. Too short sequences are dropped and too long sequences are subtracted. More details about the datasets are summarized in \tabref{tab:data}, where \#E/S denotes the average number of elements in each set, \#S/U represents the average number of sets for each user.
\begin{table}[!ht]
\centering
\normalsize
\caption{Statistics of the datasets.}
\label{tab:data}
\begin{tabular}{cccccc}
\hline
\textbf{Datasets} & \textbf{\#sets} & \textbf{\#users} & \textbf{\#elements} & \textbf{\#E/S} & \textbf{\#S/U} \\
\hline
TaFeng             & 73,355           & 9,841              & 4,935               & 5.41                          & 7.45                               \\ 
\hline
DC   & 42,905           & 9,010               & 217                & 1.52                          & 4.76                              \\ 
\hline
TaoBao   & 628,618           & 113,347               & 689
                & 1.10                          & 5.55                              \\ 
\hline
TMS   & 243,394           & 15,726               & 1,565                & 2.19                          & 15.48                              \\ 
\hline
\end{tabular}
\end{table}

\begin{table*}[!htbp]
\centering
\caption{Comparisons with different methods on Top-K performance.}
\label{tab:results}
\resizebox{\textwidth}{!}
{
\begin{tabular}{c|c|ccc|ccc|ccc|ccc}
\hline
\multirow{2}{*}{Datasets} & \multirow{2}{*}{Methods} & \multicolumn{3}{c|}{K=10}                            & \multicolumn{3}{c|}{K=20}                            & \multicolumn{3}{c|}{K=30}                            & \multicolumn{3}{c}{K=40}                            \\ \cline{3-14} 
                          &                          & \textbf{Recall}          & \textbf{NDCG}            & \textbf{PHR}             & \textbf{Recall}          & \textbf{NDCG}            & \textbf{PHR}             & \textbf{Recall}          & \textbf{NDCG}            & \textbf{PHR}             & \textbf{Recall}          & \textbf{NDCG}            & \textbf{PHR}             \\ \hline
\multirow{6}{*}{TaFeng}   & Top                      & 0.1025          & 0.0974          & 0.3047          & 0.1227          & 0.1033          & 0.3682          & 0.1446          & 0.1104          & 0.4256          & 0.1561          & 0.1140           & 0.4474          \\
                          & PersonalTop              & 0.1214          & 0.1128          & 0.3763          & 0.1675          & 0.1280           & 0.4713          & 0.1882          & 0.1336          & 0.5063          & 0.2022          & 0.1398          & 0.5292          \\
                          & ElementTransfer             & 0.0613          & 0.0644          & 0.2255          & 0.0721          & 0.0670           & 0.2519          & 0.0765          & 0.0676          & 0.2590           & 0.0799          & 0.0687          & 0.2671          \\
                          & DREAM                    & 0.1174          & 0.1047          & 0.3088          & 0.1489          & 0.1143          & 0.3814          & 0.1719          & 0.1215          & 0.4383          & 0.1885          & 0.1265          & 0.4738          \\
                          & Sets2Sets                & 0.1427          & 0.1270           & 0.4347          & 0.2109          & 0.1489          & 0.5500            & 0.2503          & 0.1616          & 0.6044          & 0.2787          & 0.1700            & 0.6379          \\
                          & DNNTSP                   & \textbf{0.1752} & \textbf{0.1517} & \textbf{0.4789} & \textbf{0.2391} & \textbf{0.1720}  & \textbf{0.5861} & \textbf{0.2719} & \textbf{0.1827} & \textbf{0.6313} & \textbf{0.2958} & \textbf{0.1903} & \textbf{0.6607} \\ \hline
\multirow{6}{*}{DC}       & Top                      & 0.1618          & 0.0880           & 0.2274          & 0.2475          & 0.1116          & 0.3289          & 0.3204          & 0.1288          & 0.4143          & 0.3940           & 0.1448          & 0.4997          \\
                          & PersonalTop              & 0.4104          & 0.3174          & 0.5031          & 0.4293          & 0.3270           & 0.5258          & 0.4499          & 0.3318          & 0.5496          & 0.4747          & 0.3332          & 0.5785          \\
                          & ElementTransfer             & 0.1930           & 0.1734          & 0.2546          & 0.2280           & 0.1816          & 0.3017          & 0.2589          & 0.1929          & 0.3417          & 0.2872          & 0.1955          & 0.3783          \\
                          & DREAM                    & 0.2857          & 0.1947          & 0.3705          & 0.3972          & 0.2260           & 0.4964          & 0.4588          & 0.2407          & 0.5613          & 0.5129          & 0.2524          & 0.6184          \\
                          & Sets2Sets                & 0.4488          & 0.3136          & 0.5458          & 0.5143          & 0.3319          & 0.6162          & 0.5499          & 0.3405          & 0.6517          & 0.6017          & 0.3516          & 0.7005          \\
                          & DNNTSP                   & \textbf{0.4615} & \textbf{0.3260}  & \textbf{0.5624} & \textbf{0.5350}  & \textbf{0.3464} & \textbf{0.6339} & \textbf{0.5839} & \textbf{0.3578} & \textbf{0.6833} & \textbf{0.6239} & \textbf{0.3665} & \textbf{0.7205} \\ \hline
\multirow{6}{*}{TaoBao}   & Top                      & 0.1567          & 0.0784          & 0.1613          & 0.2494          & 0.1019          & 0.2545          & 0.3166          & 0.1164          & 0.3220           & 0.3679          & 0.1264          & 0.3745          \\
                          & PersonalTop              & 0.2190           & 0.1535          & 0.2230           & 0.2260           & 0.1554          & 0.2306          & 0.2354          & 0.1575          & 0.2402          & 0.2433          & 0.1590           & 0.2484          \\
                          & ElementTransfer             & 0.1190           & 0.1153          & 0.1217          & 0.1253          & 0.1166          & 0.1284          & 0.1389          & 0.1197          & 0.1427          & 0.1476          & 0.1214          & 0.1516          \\
                          & DREAM                    & 0.2431          & 0.1406          & 0.2491          & 0.3416          & 0.1657          & 0.3483          & 0.4060           & 0.1796          & 0.4129          & 0.4532          & 0.1889          & 0.4606          \\
                          & Sets2Sets                & 0.2811          & 0.1495          & 0.2868          & 0.3649          & 0.1710           & 0.3713          & 0.4267          & 0.1842          & 0.4327          & 0.4672          & 0.1922          & 0.4739          \\
                          & DNNTSP                   & \textbf{0.3035} & \textbf{0.1841} & \textbf{0.3095} & \textbf{0.3811} & \textbf{0.2039} & \textbf{0.3873} & \textbf{0.4347} & \textbf{0.2154} & \textbf{0.4406} & \textbf{0.4776} & \textbf{0.2238} & \textbf{0.4843} \\ \hline
\multirow{6}{*}{TMS}      & Top                      & 0.2627          & 0.1627          & 0.4619          & 0.3902          & 0.2017          & 0.6243          & 0.4869          & 0.2269          & 0.7222          & 0.5605          & 0.2448          & 0.8007          \\
                          & PersonalTop              & 0.4508          & 0.3464          & 0.6440           & 0.5274          & 0.3721          & 0.7146          & 0.5453          & 0.3765          & 0.7339          & 0.5495          & 0.3771          & 0.7374          \\
                          & ElementTransfer             & 0.3292          & 0.2984          & 0.4752          & 0.3385          & 0.3038          & 0.4828          & 0.3410           & 0.3034          & 0.4863          & 0.3423          & 0.3036          & 0.4889          \\
                          & DREAM                    & 0.3893          & 0.3039          & 0.6090           & 0.4962          & 0.3379          & 0.7279          & 0.5677          & 0.3570           & 0.794           & 0.6155          & 0.3690           & 0.8315          \\
                          & Sets2Sets                & \textbf{0.4748} & \textbf{0.3782} & \textbf{0.6933} & 0.5601          & \textbf{0.4061} & 0.7594          & 0.6145          & \textbf{0.4204} & 0.8131          & 0.6627          & \textbf{0.4321} & 0.8570           \\
                          & DNNTSP                   & 0.4693          & 0.3473           & 0.6825          & \textbf{0.5826} & 0.3839          & \textbf{0.7880} & \textbf{0.6440} & 0.4000          & \textbf{0.8439} & \textbf{0.6840} & 0.4097          & \textbf{0.8748} \\ \hline
\end{tabular}
}
\end{table*}
\subsection{Compared Methods}
We compare our model with the following baselines, including both classical and the state-of-the-art methods:
\begin{itemize}
	\item \textbf{TOP}: 
    It uses the most popular elements appearing in the training set as the prediction for users in the test set.

    \item \textbf{PersonalTOP}: 
    It sorts the most popular elements that appear in historical sets of a given user, and then recommends them to the user as the prediction result.

    \item \textbf{ElementTransfer}: 
    ElementTransfer first learns transfer relationships between elements based on adjacent behaviors of a given user. Then it provides elements which are more likely to appear in the next-period based on the last status of the user using the learned transfer relationships.
    
    \item \textbf{DREAM}:
    This method considers both dynamic representations of users and global interactions among sets based on neural networks for next-basket recommendation \cite{yu2016dynamic}. DREAM uses max pooling operations to generate representations of baskets. Then the sequence of baskets is fed into an RNN structure which predicts the next-period basket.
    
    \item \textbf{Sets2Sets}:
    Sets2Sets \cite{DBLP:conf/kdd/HuH19} uses the average pooling operation to map sets into structured vectors and designs an encoder-decoder framework to complete multi-period sets prediction based on the attention mechanism. It also takes the repeated patterns in user behaviors into consideration. 
\end{itemize}

\subsection{Evaluation Metrics}
To get a comprehensive evaluation of the proposed method, three metrics: Recall, Normalized Discounted Cumulative Gain (NDCG) and Personal Hit Ratio (PHR) are adopted as evaluation metrics. 

Recall is widely used in estimating model performance which measures the model’s ability to find all relevant elements. For user $u_i$, Recall is calculated by
\begin{equation}
    \notag
    \mathrm{Recall@K}(u_i) = \frac{|\hat{S}_i \cap S_i|}{|S_i|},
\end{equation}
where $\hat{S}_i$ and $S_i$ are the predicted top-K elements and the ground truth of user $u_i$ respectively, $|S|$ denotes the size of set $S$. We adopt the average recall of all users as a metric.

NDCG is a measure of ranking quality which considers the order of elements in a list. For user $u_i$, NDCG is calculated by
\begin{equation}
    \notag
    \mathrm{NDCG@K}(u_i) = \frac{\sum_{k = 1}^{K}{\frac{\delta(\hat{S}_i^k, S_i)}{\log_2(k + 1)}}}{\sum_{k = 1}^{\min(K, |S_i|)}{\frac{1}{\log_2\left(k + 1\right)}}},
\end{equation}
where $\delta\left(v,S\right)$ returns 1 when the element $v$ is in the set $S$, otherwise 0. The average NDCG of all users is adopted as another metric. 

PHR pays attention to the performance at user level which represents the ratio of users whose predicted sets contain the elements appearing in the ground truth. PHR is calculated by
\begin{equation}
    \notag
    \mathrm{PHR@K} = \frac{\sum_{i=1}^{N^\prime} {\mathds{1}\left(|\hat{S}_i \cap S_i|\right)}}{N^\prime},
\end{equation}
where $N^\prime$ denotes the number of testing samples, $\mathds{1}\left(x\right)$ is an indicator function which returns 1 when $x\geq0$, otherwise 0.

\subsection{Experimental Settings}
For evaluation, we generate a ranking list of top-K elements from the output and K is set to 10, 20, 30 and 40 respectively. We divide each dataset into train, validation and test set across users with the ratios of 70\%, 10\% and 20\% to do experiments. After partitioning the data, we train our model on the training data for a fix number of epochs (e.g. 100 epochs), and choose the model which achieves the best performance on the validation set for testing. Adam \cite{kingma2014adam} is adopted as the optimizer in our experiment. We utilize batch normalization \cite{DBLP:conf/icml/IoffeS15} technique between weighted convolutional layers to accelerate the training speed.
The learning rate on TaFeng, TaoBao and TMS datasets is set to 0.001, and it is set to 0.005 on DC dataset.
We stack 2 weighted convolutional layers and employ 4 attention heads on all four datasets. The hidden dimension $F$ and batch size are set to 32 and 64 respectively. The model is implemented with the PyTorch framework. We make the code and data publicly available on GitHub platform \footnote{Code and data are available at https://github.com/yule-BUAA/DNNTSP.}.

\subsection{Experimental Results and Analysis}
The comparisons of DNNTSP with other methods on Top-K performance are reported in \tabref{tab:results}. By analyzing the results, some conclusions could be summarized.

\begin{figure*}[!htbp]
    \centering
    \includegraphics[width=2\columnwidth]{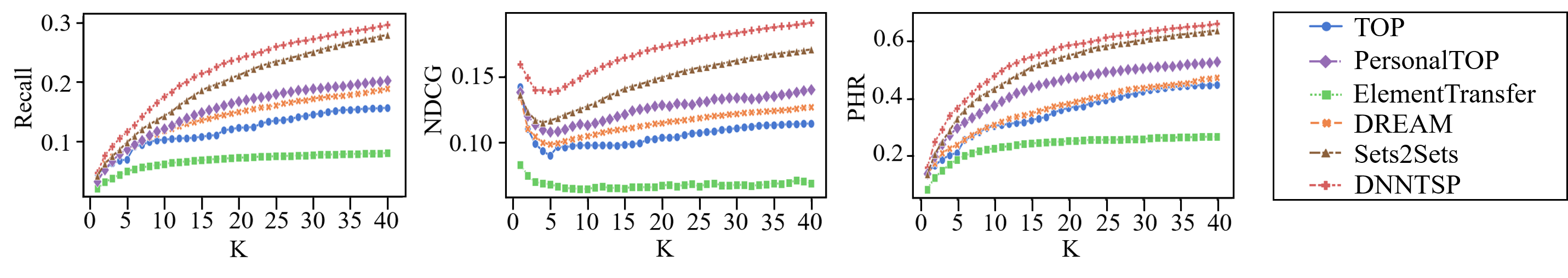}
    \caption{The performance of DNNTSP on different values of top-K on TaFeng dataset.}
    \label{fig:different_TOPK_range}
\end{figure*}

\begin{figure}[!htbp]
    \centering
    \includegraphics[width=0.96\columnwidth]{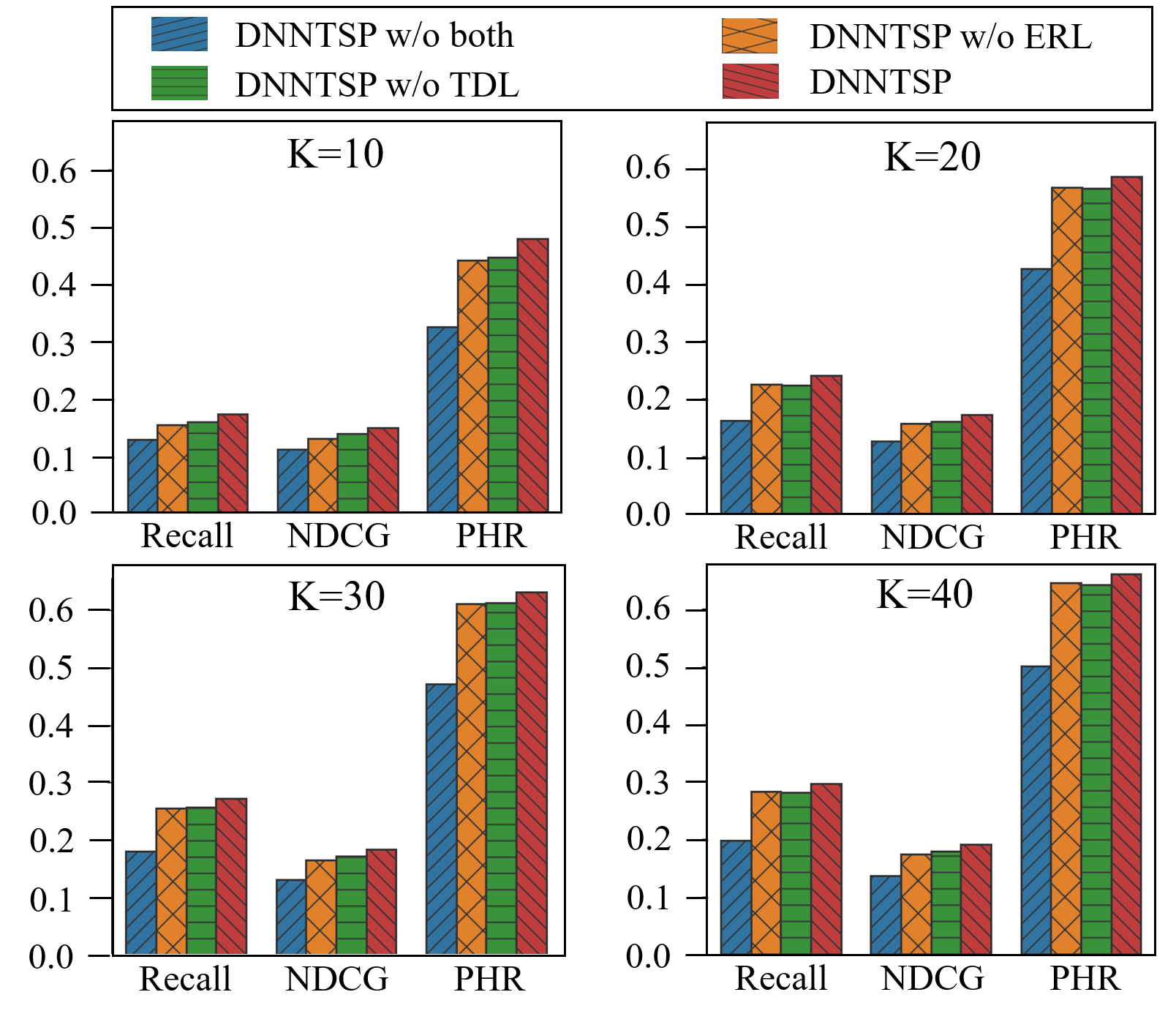}
    \caption{Effects of the ERL and TDL components on TaFeng dataset, and K is set to 10, 20, 30 and 40 respectively.}
    \label{fig:ablation_study_TaFeng}
\end{figure}

\begin{figure*}[!htbp]
    \centering
    \includegraphics[width=2\columnwidth]{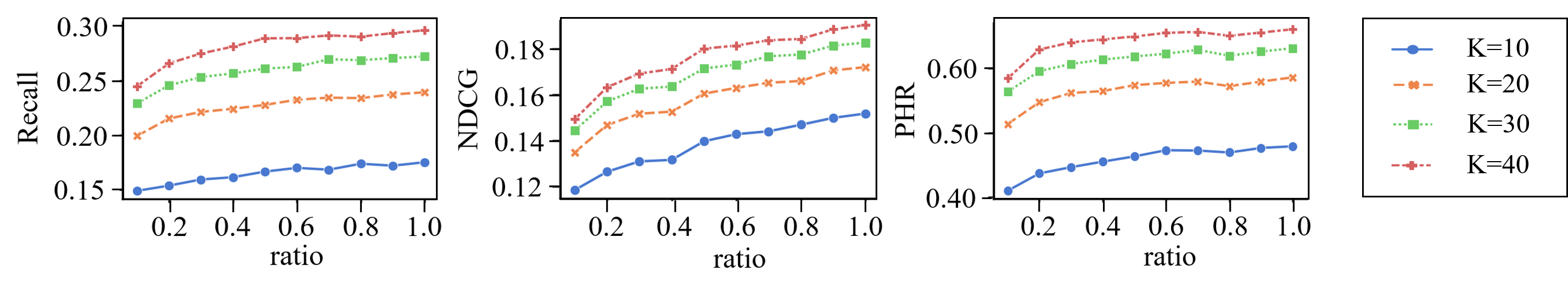}
    \caption{The performance of DNNTSP on different ratios of the training data on TaFeng dataset.}
    \label{fig:different_ratios}
\end{figure*}

Firstly, PersonalTOP achieves competitive or even better performance than other baselines in some cases although it does not consider the temporal dependency. 
This is because that users tend to interact with some elements repeatedly due to their preferences, which are not affected by the time.
PersonalTOP performs better than TOP because it could provide personalized results for different users. TOP gets worse metrics as it always provides the same elements for all users.

Secondly, ElementTransfer performs worse than DREAM as it only considers adjacent temporal dependency, while DREAM focuses on the whole sequence due to RNN structures. It indicates that users' behaviors are temporally dependent. So learning the temporal dependency from the whole sequence of the users could obtain better prediction performance.

Thirdly, Sets2Sets achieves better performance than other baselines in most cases because it learns temporal dependency by RNNs combined with the attention mechanism, which helps to select the most useful temporal dependencies in the sequence. What's more, it considers the frequent behaviors of users by modelling the repeated patterns, which also improves the prediction performance.

Finally, the proposed DNNTSP outperforms all other methods significantly in most cases. Compared with TOP and PersonalTOP, DNNTSP learns dynamic temporal dependency in users' sequential behaviors. Compared with ElementTransfer and DREAM, DNNTSP focuses on the temporal dependency of the whole sequence and leverages attention mechanism to adaptively select the most important temporal dependencies. Compared with Sets2Sets, DNNTSP learns set-level element relationship, which could maintain useful information as much as possible. What's more, DNNTSP learns the interactions of elements from a global perspective by mining shared patterns in different sequences. We also observe that DNNTSP could not outperform Sets2Sets completely in TMS dataset, especially on the NDCG metric. We infer that the repeated behaviors in TMS dataset are more obvious than that in other datasets, which result in a higher ranking quality. We will investigate this phenomenon in a further step in \secref{section-4-repeated-behaviors}. 

In order to compare our method with baselines more comprehensively, we also show the performance of the proposed model when top-K varies in consecutive values. Due to space limitations, we just show the results on TaFeng dataset. As shown in \figref{fig:different_TOPK_range}, we can see that DNNTSP outperforms other baselines consistently when the value of K changes from 1 to 40. This indicates that our method could provide more precise predictions without the influence of the capacity of predicted sets.

\begin{table*}[!htbp]
\centering
\caption{Effects of the repeated behaviors modelling component on TMS dataset.}
\label{tab:frequency_results}
\resizebox{\textwidth}{!}
{
\begin{tabular}{c|c|ccc|ccc|ccc|ccc}
\hline
\multirow{2}{*}{Dataset} & \multirow{2}{*}{Methods} & \multicolumn{3}{c|}{K=10}                            & \multicolumn{3}{c|}{K=20}                            & \multicolumn{3}{c|}{K=30}                            & \multicolumn{3}{c}{K=40}                            \\ \cline{3-14} 
                          &                          & \textbf{Recall}          & \textbf{NDCG}            & \textbf{PHR}             & \textbf{Recall}          & \textbf{NDCG}            & \textbf{PHR}             & \textbf{Recall}          & \textbf{NDCG}            & \textbf{PHR}             & \textbf{Recall}          & \textbf{NDCG}            & \textbf{PHR}             \\ \hline
\multirow{4}{*}{TMS}    & Sets2Sets-                & 0.3954          & 0.3494          & 0.6198          & 0.4845          & 0.3771          & 0.7216          & 0.5539          & 0.3956          & 0.7943          & 0.5975          & 0.4062          & 0.8328           \\
& Sets2Sets                & 0.4748 & 0.3782 & 0.6933 & 0.5601          & 0.4061 & 0.7594          & 0.6145          & 0.4204 & 0.8131          & 0.6627          & 0.4321 & 0.8570           \\
                          & DNNTSP                   & 0.4693          & 0.3473           & 0.6825          & 0.5826 & 0.3839          & 0.7880 & 0.6440 & 0.4000          & 0.8439 & 0.6840 & 0.4097          & 0.8748 \\
                          & DNNTSP+                   & \textbf{0.4883} & \textbf{0.3805} & \textbf{0.7092} & \textbf{0.6066} & \textbf{0.4179} & \textbf{0.8086} & \textbf{0.6684} & \textbf{0.4343} & \textbf{0.8665} & \textbf{0.7061} & \textbf{0.4435} & \textbf{0.8922} \\ \hline
\end{tabular}
}
\end{table*}

\subsection{Ablation Study}
To investigate the effects of element relationship learning and temporal dependency learning components, we conduct the ablation study by removing the two components manually and comparing the performance with the original DNNTSP. 

Specifically, three-fold ablation experiments have been implemented: 1) We remove the Element Relationship Learning component by stacking the representations of appearing elements in the sequence based on the sequence's length (denoted as DNNTSP w/o ERL), which means that we ignore the relationships between elements and and do not propagate information among elements. 
2) We replace the Temporal Dependency Learning component by simply aggregating the sequence of each appearing element using average pooling (denoted as DNNTSP w/o TDL), which means that we do not consider the evolutionary pattern in the sequence and lose some temporal information. 
3) Moreover, we remove both the two components simultaneously (denoted as DNNTSP w/o both) to conduct experiments. 
Experimental results on TaFeng dataset are shown in \figref{fig:ablation_study_TaFeng}.

From the results, we could conclude that the performance of DNNTSP decreases when any component is abandoned, because the ERL component takes element relationship into consideration and the TDL module learns dynamic temporal dependency from the whole sequence and selects the most important dependencies adaptively. DNNTSP w/o ERL ignores the element relationship and DNNTSP w/o TDL omits the evolution of dynamic changes in the sequence, so they both obtain worse performance. DNNTSP (w/o both) achieves the worst results as it does not consider either the element relationship or temporal dependency in the sequence.

\subsection{Effects of the Training Data Ratio}
To demonstrate the effectiveness of the gated information fusing component, we train our model on training set with varying sizes. Specifically, we randomly choose data in the original training set by changing the ratio from 10\% to 100\% with 10\% increment each time. Finally, we could generate 10 training sets with different sizes and train the model on each dataset.

Experimental results on TaFeng dataset are shown in \figref{fig:different_ratios}.
From the results, we could observe that our model performs better when the size of training data increases. More importantly, our model is able to achieve competitive performance when it is trained with only forty percent of the training data. This proves that the gated information fusing component helps our model discover the shared patterns in different sequences, and therefore our model could get satisfactory results with only a portion of the training data. The results illustrate that our model is applicable to the scenarios with sparse data.

\subsection{Effects of Modelling the Repeated Behaviors in Temporal Sets Prediction}\label{section-4-repeated-behaviors}
Since our model could not outperform Sets2Sets in some metrics on the TMS dataset, we conclude that the repeated behaviors have a greater impact on the TMS dataset and the component of modelling such behaviors in Sets2Sets helps Sets2Sets achieve better performance. So we study the effects of modelling the repeated behaviors in temporal sets prediction and use the TMS dataset to conduct experiments. Specifically, we first remove the repeated behaviors modelling component in Sets2Sets and denote the model as Sets2Sets-. Then we incorporate this component into our DNNTSP and denote it as DNNTSP+. Experimental results of the modified models are shown in \tabref{tab:frequency_results}.

From the results, we could conclude that in the same condition, the proposed DNNTSP performs better than Sets2Sets. 
On the one hand, without modelling repeated behaviors, DNNTSP achieves better results than Sets2Sets-, which shows the superiority of DNNTSP over Sets2Sets in temporal sets prediction when no empirical information is added. 
On the other hand, DNNTSP+ also outperforms Sets2Sets, which proves the effectiveness of modelling the repeated behaviors in temporal sets prediction and also demonstrates that our model is able to achieve the best performance by incorporating the repeated behaviors modelling component.

\section{Related work}\label{section-5}
This section reviews the existing literature related to our work, and also points out the differences of previous studies with our research.


\textbf{Next-period Set Prediction}.
In the field of retail, \citet{yu2016dynamic} used pooling operations among products in each basket to get its representation and employed an RNN structure to learn the dynamic evolves in the sequence of customer's behaviors. 
In field of health care, \citet{choi2016multi} focused more on the relationships of drugs and introduced a variant of Skip-gram model to learn drugs' co-occurrence information. Based on the learned relationships, they generated representations of prescriptions and used a softmax classifier to predict subsequent prescriptions within a context window. 
More generally, \citet{DBLP:conf/kdd/Benson0T18} studied the repeated behaviors in sequences of sets and provided a stochastic model to mine the hidden patterns. However, their model assumed that only the repeated elements would appear in next-period set and the model became prohibitive when the size of set gets larger.
\citet{DBLP:conf/kdd/HuH19} obtained the representations of sets by pooling operations and proposed an encoder-decoder framework to make multi-period sets prediction. Moreover, they considered the repeated user behaviors to improve the model performance. We could find that most of the existing methods first embedded sets into latent vectors and then predicted future sets based on the sequences of embedded sets. However, the two-step learning process usually leads to the loss of elements' information, so we provide a new perspective to deal with the temporal sets prediction problem in this paper.

\textbf{Relationship Learning Based on Graph Neural Networks}. 
Graph Network Networks (GNNs) have shown the effectiveness in learning representations with consideration of multiple complex relationships. 
GNNs first propagate information among each node and its neighbours, and then produce a representation for each node in the relationship graph based on received information \cite{DBLP:journals/corr/abs-1812-08434}.
According to different convolutional operations on the graphs, GNNs could be divided into spectral-based methods \cite{DBLP:conf/iclr/KipfW17} and spatial-based methods \cite{DBLP:conf/nips/HamiltonYL17}. 
In the studied problem, the size-variant characteristic makes sets arbitrary-sized, so we design a weighted graph convolutional network to deal with dynamic graphs. 

\textbf{Temporal Dependency Learning Based on the Attention Mechanism}.
Since the proposition of the attention mechanism in neural networks, it has achieved great success in various tasks such as image caption \cite{DBLP:conf/icml/XuBKCCSZB15} and machine translation \cite{DBLP:journals/corr/BahdanauCB14}. Inspired by the fact that people usually pay much attention on the important part of the whole perception space, attention mechanisms provide neural networks with the ability to assign larger weights on the most useful parts of the collected information.
Recently, a novel framework based solely on attention mechanism, namely Transformer, has been proposed to apply in sequential tasks successfully without using any recurrent or convolutional architectures \cite{vaswani2017attention}. Since the self-attention mechanism has a strong ability to capture both short and long-term dependency by allowing the model to access any part of historical records without the constraint of distance, we extend the self-attention mechanism in our model to learn dynamic temporal dependency in different sequences.

\section{Conclusion} \label{section-6}
This paper studies predictive modelling of a new type of temporal data, namely, temporal sets. Different from the existing methods, which adopt a set embedding procedure to turn the temporal sets prediction problem into a conventional prediction problem, our method is founded on the multiple and comprehensive set-level element representations. In particular, our method consists of three components: 1) an element relationship learning module to capture multiple set-level element relationships; 2) an attention-based temporal dependency learning module to learn the temporal dependencies of the sets and elements from the whole sequence; and 3) a gated information fusing module to discover the shared patterns among different sequences and fuse both the static and dynamic information. Experimental results demonstrate that our method could circumvent the information loss problem suffered by the set-embedding based methods, and achieve higher prediction performance than the state-of-the-art methods.

\begin{acks}
The authors would like to thank the anonymous reviewers for their constructive comments on this research work. 
This work is supported by the National Natural Science Foundation of China (51778033, 51822802, 51991395, 71901011, U1811463), the Science and Technology Major Project of Beijing (Z191100002519012) and the National Key R$\And$D Program of China (2018YFB2101003).
\end{acks}

\bibliographystyle{ACM-Reference-Format}
\bibliography{reference}

\end{document}